\documentclass{article}

\usepackage{microtype}
\usepackage{graphicx}
\usepackage{subfigure}
\usepackage{booktabs} 

\usepackage{hyperref}

\usepackage[accepted]{icml2023}

\usepackage{amsmath}
\usepackage{amssymb}
\usepackage{mathtools}
\usepackage{amsthm}
\usepackage{bm}

\usepackage[capitalize,noabbrev]{cleveref}

\usepackage{nicefrac}

\theoremstyle{plain}

\theoremstyle{definition}

\theoremstyle{remark}

\usepackage[textsize=tiny]{todonotes}

\usepackage{layouts}

\usepackage{amsmath,amsfonts,bm}

\def\eqref#1{equation~\ref{#1}}

\def\1{\bm{1}}

\def\rvw{{\mathbf{w}}}
\def\rvx{{\mathbf{x}}}
\def\rvy{{\mathbf{y}}}

\DeclareMathAlphabet{\mathsfit}{\encodingdefault}{\sfdefault}{m}{sl}
\SetMathAlphabet{\mathsfit}{bold}{\encodingdefault}{\sfdefault}{bx}{n}

\def\gW{{\mathcal{W}}}
\def\gX{{\mathcal{X}}}
\def\gY{{\mathcal{Y}}}

\newcommand{\E}{\mathbb{E}}

\newcommand{\KL}{D_{\mathrm{KL}}}

\usepackage{mathtools}
\usepackage{amsthm}
\usepackage{pifont}

\DeclarePairedDelimiterX{\infdivx}[2]{[}{]}{%
  #1\delimsize\| #2%
}
\DeclarePairedDelimiterX{\infdivxcolon}[2]{[}{]}{%
  #1\delimsize: #2%
}
\newcommand{\KLD}{\KL\infdivx}

\DeclarePairedDelimiterX{\innerProd}[2]{\langle}{\rangle}{%
    #1,#2%
}

\newcommand{\Reals}{\mathbb{R}}

\newcommand{\Normal}{\mathcal{N}}
\newcommand{\Data}{\mathcal{D}}

\icmltitlerunning{Minimal Random Code Learning with Mean-KL Parameterization}

\begin{document}

\twocolumn[
\icmltitle{Minimal Random Code Learning with Mean-KL Parameterization}



\icmlsetsymbol{equal}{*}

\begin{icmlauthorlist}
\icmlauthor{Jihao Andreas Lin}{cam}
\icmlauthor{Gergely Flamich}{cam}
\icmlauthor{José Miguel Hernández-Lobato}{cam}
\end{icmlauthorlist}

\icmlaffiliation{cam}{University of Cambridge}

\icmlcorrespondingauthor{Jihao Andreas Lin}{jal232@cam.ac.uk}

\icmlkeywords{Neural Compression}

\vskip 0.3in
]



\printAffiliationsAndNotice{}  

\begin{abstract}
This paper studies the qualitative behavior and robustness of two variants of \emph{Minimal Random Code Learning} (MIRACLE) used to compress variational Bayesian neural networks. 
MIRACLE implements a powerful, \emph{conditionally Gaussian} variational approximation for the weight posterior $Q_\rvw$ and uses relative entropy coding to compress a weight sample from the posterior using a Gaussian coding distribution $P_\rvw$. 
To achieve the desired compression rate, $\KLD{Q_\rvw}{P_\rvw}$ must be constrained, which requires a computationally expensive annealing procedure under the conventional mean-variance (Mean-Var) parameterization for $Q_\rvw$. 
Instead, we parameterize $Q_\rvw$ by its mean and KL divergence from $P_\rvw$ to constrain the compression cost to the desired value by construction. We demonstrate that variational training with Mean-KL parameterization converges twice as fast and maintains predictive performance after compression.
Furthermore, we show that Mean-KL leads to more meaningful variational distributions with heavier tails and compressed weight samples which are more robust to pruning.
\end{abstract}

\begin{figure*}[t]
\centering
\includegraphics[width=\textwidth]{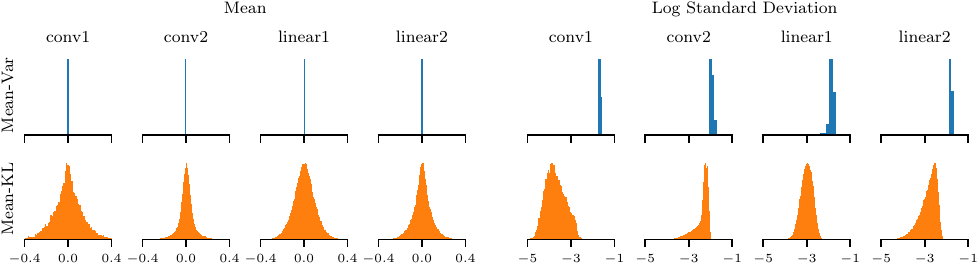}
\vspace{-20pt}
\caption{Layerwise histograms of variational mean and log standard deviation for Mean-Var (blue) versus Mean-KL (orange) parameterizations.
Mean-Var struggles to learn meaningful distributions: means are concentrated at zero and standard deviations are clustered at high values.
Mean-KL learns more reasonable distributions with heavier tails and a broader range of values.}
\label{fig:histograms}
\vspace{-10pt}
\end{figure*}

\section{Introduction}
\label{sec:introduction}
\noindent
With the ever-growing size of neural network architectures, such as large language models \citep[e.g.\ BERT,][]{kenton2019bert}, it is now a key challenge to ensure their memory and energy efficiency.
While there is a large literature on model compression, almost all works rely on some form of quantization scheme.
In this paper, we consider an alternative method to quantization, namely Minimal Random Code Learning \citep[MIRACLE,][]{Havasi19}, which has recently demonstrated state-of-the-art performance for neural network compression.
The MIRACLE framework employs a powerful, conditionally Gaussian variational distribution $Q_\rvw$ over the weights $\rvw$ of a neural network and uses relative entropy coding \citep[REC,][]{flamich2020compressing} with a Gaussian coding distribution $P_\rvw$ to encode a random weight sample from $Q_\rvw$.
The average coding cost of encoding a weight sample is $\KLD{Q_\rvw}{P_\rvw}$, which needs to be carefully controlled in a practical compression scheme.
To this end, we propose to use Mean-KL parameterization for Gaussians \citep{Flamich22} to parameterize $Q_\rvw$, allowing explicit control over $\KLD{Q_\rvw}{P_\rvw}$ by construction.
We demonstrate that Mean-KL leads to many practical benefits over the conventional mean-variance (Mean-Var) parameterization used by \citealt{Havasi19}, which requires a computationally expensive annealing procedure to control the coding cost.
In particular, we show that, compared to Mean-Var parameterization, variational training converges in half the number of iterations using Mean-KL parameterization while maintaining predictive performance after compression.
Furthermore, we illustrate that the resulting variational distribution exhibits more meaningful shapes with heavy tails, which makes the compressed weight sample more robust against zero pruning.

\section{Background}
\label{sec:background}
\noindent
\paragraph{Minimal Random Code Learning}
\citealt{Havasi19} consider a setting akin to the $\beta$-VAE \citep{Higgins2016betaVAELB} to encode neural network weights with a limited information budget $C$.
To this end, let $\gX, \gY$ and $\gW$ be the input, output and weight spaces, respectively, let $\Data = \{ (\rvx_n, \rvy_n)\}_{n=1}^N$ be a dataset and let $h: \gX \times \gW \to \gY$ be a neural network with input $\rvx$ and weights $\rvw$.
To control the information content of the weights, let $P_\rvw$ be the \emph{coding distribution} and $Q_\rvw$ be the \emph{variational distribution} over $\rvw$.
In this setting, \citealt{hinton1993keeping} show that the information content of the weights is $\KLD{Q_\rvw}{P_\rvw}$.
Further, let ${\Delta: \mathcal{Y} \times \mathcal{Y} \to \Reals^+}$ be a \emph{distortion function}.
MIRACLE minimizes
\begin{align}
    \E_{\rvw \sim Q_\rvw} \sum_{(\rvx, \rvy) \in \Data} \Delta(\rvy, h(\rvx, \rvw))
    + \beta\KLD{Q_\rvw}{P_\rvw}
    \label{eq:beta_elbo}
\end{align}
with respect to $Q_\rvw$ to minimize distortion within the given information budget of $\KLD{Q_\rvw}{P_\rvw} = C$ nats.
During optimization, $\beta$ is dynamically adapted to anneal the KL divergence, such that the constraint is eventually satisfied.
\par
In this paper, we encode the samples using minimal random coding \citep[MRC,][]{Havasi19} for simplicity, though more sophisticated approaches, such as A* coding \citep{Flamich22} or greedy Poisson rejection sampling \citep{flamich2023greedy}, have been invented.
Given a suitable $Q_\rvw$, a random sample from $Q_\rvw$ is compressed by first drawing $K = \exp(\KLD{Q_\rvw}{P_\rvw})$ samples from $P_\rvw$.
These $K$ samples are then used to construct a discrete distribution whose probability mass function is defined by the importance weights $r_k = \frac{dQ_\rvw}{dP_\rvw}(\rvw_k)$, where $\frac{dQ_\rvw}{dP_\rvw}$ is the Radon-Nikodym derivative, i.e. the density ratio, of $Q_\rvw$ with respect to $P_\rvw$.
The compressed weight sample is represented by an index $k_* \sim Q_k$.
Since $0 \leq k_* < K$, it is always possible to encode $k_*$ using $\KLD{Q_\rvw}{P_\rvw} = C$ nats.
The weight sample can be decoded by drawing the $k_*$\textsuperscript{th} sample from $P_\rvw$ using a shared random number generator with a shared random seed.
Due to the exponential scaling, simulating $K$ samples is intractable if $\rvw$ has many dimensions.
\citealt{Havasi19} solve this issue by partitioning $\rvw$ dimensionwise into smaller blocks with local information budgets $C_\mathrm{block}$, such that $K$ is feasible.
\par
\paragraph{Refining Mean-Field Posteriors}
An important choice in practice is the variational family over which we optimize \cref{eq:beta_elbo}.
Since we are interested in studying the behavior of samples using MIRACLE, we also adopt the variational family suggested by \citet{Havasi19}.
Concretely, assume that we have already partitioned the weight vector as $\rvw = w_{1:B} = \rvw_1 \oplus \rvw_2 \oplus \hdots \oplus \rvw_B$, where $B$ denotes the number of blocks, and $\oplus$ denotes vector concatenation.
To begin, we use a mean-field Gaussian variational approximation, i.e.\ we parameterize the means $\mu_{1:B} = \mu_1 \oplus \hdots  \oplus \mu_B$ and marginal variances $\sigma^2_{1:B} = \sigma^2_1 \oplus \hdots \oplus \sigma^2_{B}$ (Mean-Var).
Once variational training converges, we compress the first block $\rvw_1$, resulting in a sample $\tilde{\rvw}_1$.
Keeping $\tilde{\rvw}_1$ fixed, we resume optimization to \textit{fine-tune} the remaining means $\mu_{2:B}$ and variances $\sigma^2_{2:B}$.
We repeat this process $B$ times in total, where at step $b$, $\tilde{\rvw}_1, \hdots, \tilde{\rvw}_{b-1}$ are fixed, means $\mu_{b:B}$ and variances $\sigma^2_{b:B}$ are optimized, and a random sample from block $b$ is encoded. 
Note that the variational posterior $Q_{\rvw_{b:B} | \tilde{\rvw}_{1:b - 1}}$ at step $b$ is only \textit{factorized conditionally} on the weight samples in the first $b - 1$ blocks, which results in a much better variational approximation.
\par
\paragraph{Mean-KL Parameterization for Gaussians}
\citealt{Flamich22} show that, given a univariate Gaussian coding distribution $P_w = \Normal(w|\nu, \rho^2)$ with mean $\nu$ and variance $\rho^2$, a variational distribution $Q_w = \Normal(w|\mu, \sigma^2)$ can be uniquely parameterized by mean $\mu$ and $\KLD{Q_w}{P_w} = \kappa$ if
\begin{align}
\label{eq:mean_constraint}
    |\mu - \nu| < \rho \sqrt{2 \kappa}
\end{align}
is satisfied.
The variance $\sigma^2$ of $Q_w$ can be recovered via
\begin{align}
\label{eq:mean_kl_variance}
    \sigma^2 = - \rho^2 W\left(-\exp(z^2 - 2\kappa - 1) \right),
\end{align}
where $z = (\mu - \nu) / \rho$ and $W$ is the principal branch of the Lambert $W$ function \citep{corless1996lambert}, defined by the relation $W(x)e^{W(x)} = x$ (see \Cref{app:lambert_w} for details).
\section{Mean-KL Parameterization for MIRACLE}
\label{sec:method}
Recognizing that the main goal of minimizing \Cref{eq:beta_elbo} combined with KL annealing is to solve
\begin{align}
    \underset{Q_\rvw}{\arg \min} \quad \E_{\rvw \sim Q_\rvw} \sum_{(\rvx, \rvy) \in \Data} \Delta(\rvy, h(\rvx, \rvw)), \\
    \text{subject to} \quad \KLD{Q_\rvw}{P_\rvw} = C,
\end{align}
we propose to use Mean-KL parameterization \cite{Flamich22} to enforce the $\KLD{Q_\rvw}{P_\rvw} = C$ constraint mathematically instead of performing computationally expensive KL annealing.
To this end, the total information budget $C = \kappa$ must be distributed to each weight, resulting in local information budgets $\kappa_w$.
Thus, in Mean-KL parameterization, each weight has a mean parameter $\mu_w$ and a local information budget $\kappa_w$, matching the number of parameters for the conventional Mean-Var parameterization, albeit with one fewer degree of freedom because $\sum_{w \in \rvw} \kappa_w = \kappa$.

In practice, we introduce an \emph{information quota} parameter $\gamma_w$ per weight, which satisfies $\sum_{w \in \rvw} \gamma_w = 1$ and defines the relative share of the total information budget assigned to $w$, that is $\kappa_w = \gamma_w \kappa$.
The constraint on the information quota parameters is implemented using a softmax function.
To ensure that $|\mu_w - \nu| < \rho \sqrt{2 \kappa_w}$ (\Cref{eq:mean_constraint}),
we define
\begin{align}
\label{eq:mean_kl_mean}
    \mu_w = \nu + \rho \sqrt{2\kappa_w} \mathrm{tanh}(\tau_w),
\end{align}
as suggested by \citet{Flamich22}, leaving $\tau_w$ and $\gamma_w$ as trainable parameters.
In combination with blockwise partitioning of $\rvw$, each block has its own constraint and $\kappa$ is simply replaced by $\kappa_{\mathrm{block}}$.
When drawing samples from $Q_\rvw$ or evaluating the density of $Q_\rvw$, we convert $\tau_w$ and $\gamma_w$ to $\mu_w$ and ${\sigma_w}^2$ using \Cref{eq:mean_kl_mean} and \Cref{eq:mean_kl_variance}, respectively, followed by the same computations as with conventional Mean-Var parameterization.

\section{Experiments}
\label{sec:experiments}
We empirically demonstrate advantages of Mean-KL compared to conventional Mean-Var parameterization:
We show that variational training with Mean-KL parameterization converges faster than Mean-Var while maintaining predictive performance, we illustrate that Mean-KL leads to more meaningful distributions with heavier tails, and we demonstrate that these more meaningful distributions translate to improved robustness when pruning weights to zero.

\paragraph{Training Dynamics and Predictive Performance}
We adopt the experimental setup of \citealt{Havasi19} and train a LeNet-5 on MNIST.
The distortion function $\Delta$ is the cross-entropy, which is commonly used as a loss function in image classification.
Matching \citealt{Havasi19}, we used a local information budget of $C_{\mathrm{block}} = \kappa_{\mathrm{block}} = 20$ bits.
We varied the block size between 20, 30, and 40.
For both parameterizations, we used Adam with a learning rate of 0.001 and a mini-batch size of 200.
For KL divergence annealing with Mean-Var, we used $\epsilon_{\beta_0} = 10^{-8}$ and $\epsilon_{\beta} = 5 \times 10^{-5}$, as suggested by \citealt{Havasi19}.
See \Cref{app:implementation_details} for further implementation details.

\begin{figure}[t]
\centering
\includegraphics[width=.48\textwidth]{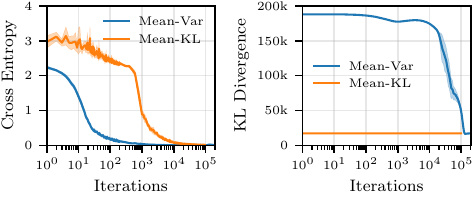}
\vspace{-20pt}
\caption{Training dynamics of Mean-Var and Mean-KL parameterizations.
Mean-Var requires a large amount of iterations to anneal the KL divergence to the desired coding cost.
Mean-KL constrains $\KLD{Q_\rvw}{P_\rvw}$ to the desired value and focuses on minimizing cross entropy, converging in half the number of iterations.}
\label{fig:optim_trace}
\vspace{-10pt}
\end{figure}

Figure~\ref{fig:optim_trace} illustrates how Mean-Var spends most of the optimization on minimizing and annealing the KL divergence to the desired coding cost, whereas for Mean-KL, the whole optimization process focuses on minimizing cross entropy, given that the parameterization already constrains the KL divergence to the desired coding cost.
Crucially, KL divergence annealing with Mean-Var takes a tremendous amount of time while minimizing cross entropy with Mean-KL converges in just \emph{half} the number of iterations.
Table~\ref{tab:mnist} shows that Mean-KL maintains predictive performance comparable to Mean-Var across different compression ratios, being slightly better in the low compression ratio setting and slightly worse in the high compression ratio settings, albeit within standard error.
\begin{table}[t]
\centering
\caption{MNIST classification error after compression (lower is better). Mean $\pm$ standard error over 10 seeds.}
\label{tab:mnist}
\vspace{5pt}
\begin{tabular}{c c c c}
\toprule
Block Size & Ratio & Mean-Var & Mean-KL \\
\midrule
20 & 555x & $0.82 \pm 0.07$ \% & $0.77 \pm 0.05$ \% \\
30 & 833x & $0.79 \pm 0.05$ \% & $0.87 \pm 0.08$ \% \\
40 & 1111x & $0.87 \pm 0.07$ \% & $0.96 \pm 0.08$ \% \\
\midrule
\multicolumn{2}{c}{Optimizer Iterations} & 200,000 & 100,000 \\
\bottomrule
\end{tabular}
\end{table}

\paragraph{Visualizing Variational Posteriors}
\begin{figure*}[t]
\centering
\includegraphics[width=\textwidth]{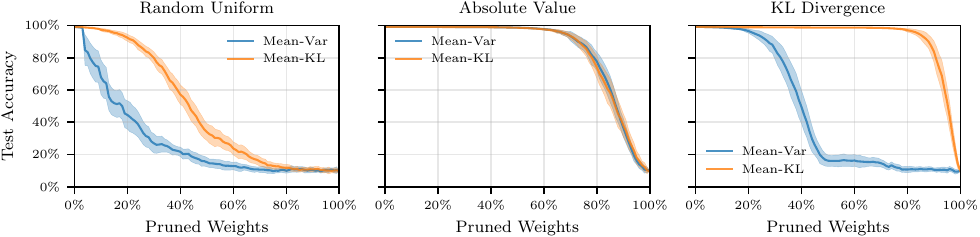}
\vspace{-20pt}
\caption{Predictive performance of compressed weight samples from Mean-Var and Mean-KL parameterizations when exposed to pruning via setting weights to zero by selecting the pruned weights uniformly at random (left), based on the smallest absolute values (middle) or based on minimizing KL divergence to a Dirac delta centered at zero (right).
Mean $\pm$ standard error over block sizes 20, 30, and 40 with 10 random seeds per block size.}
\label{fig:pruning}
\vspace{-10pt}
\end{figure*}

To qualitatively investigate the variational posterior distributions, we plot layerwise histograms of learned parameters after the compressed weight sample has been generated.
For purposes of comparison, both Mean-Var and Mean-KL parameters have been have been converted to mean and log standard deviation.

Figure~\ref{fig:histograms} reveals striking differences between layerwise Mean-Var and Mean-KL parameter distributions.
In terms of the means, Mean-Var parameters collapse to sharp peaks at zero for all layers without any visible tails.
In contrast, Mean-KL mean parameters manifest much wider, symmetric distributions centered around zero with heavier tails, resembling shapes akin to Laplace, Gaussian or Student's $t$-distributions.
In terms of the log standard deviation, similarly, Mean-Var parameters form peaked distributions around a particular value with virtually no tails.
The distributions of Mean-KL log standard deviations is more spread out, forming distinct shapes for each layer.
In general, Mean-Var standard deviations seem to be higher than Mean-KL standard deviations.
Furthermore, despite resulting in similar predictive performance, the stark differences in distributional shapes suggest potential qualitative differences between the learned variational posteriors.

\paragraph{Robustness to Pruning}
To study potential qualitative differences between variational posteriors learned using Mean-Var and Mean-KL parameterizations, we analyze the robustness of the compressed weight sample by setting certain weights to zero using three different strategies:
\begin{enumerate}
    \item Random Uniform: Select pruned weights uniformly at random.
    This strategy reflects a general notion of robustness due to the uninformed nature of this strategy.
    \item Absolute Value: Set the weight with smallest absolute value to zero.
    This strategy is a simple yet competitive pruning baseline \cite{blalock2020state}, which only depends on the compressed weight sample itself.
    If the same sample was generated by two different distributions it would still be pruned in the same way.
    \item KL Divergence: Prune the weight which minimizes the KL divergence from the variational posterior to a Dirac delta at zero, ${\arg \min}_i \, \KLD{\delta_w}{Q_{w_i}}$.
    For a Gaussian variational posterior with diagonal covariance matrix, this is equivalent to finding the weight with maximal density at zero (see Appendix~\ref{app:kl_pruning} for details).
    This strategy depends on the variational posterior, implying that the same compressed sample would be pruned differently if it was generated by two different distributions.
\end{enumerate}
Figure~\ref{fig:pruning} illustrates how the test accuracy changes as more weights in the compressed sample are pruned to zero.
With Random Uniform pruning, Mean-Var test accuracy quickly drops off, already losing more than half the performance after about 20\% of the weights have been pruned, and diminishing to performance equal to guessing uniformly at random after roughly 70\% of the weights have been set to zero.
Mean-KL performance also reduces rapidly, albeit more gracefully.
After setting 30\% of all weights to zero, a test accuracy of 80\% is maintained.
Performance equal to guessing is reached after more than 80\% of the weights have been pruned.
This suggests a general notion of improved robustness of the compressed sample produced by Mean-KL compared to Mean-Var.

With Absolute Value pruning, Mean-Var and Mean-KL perform nearly identical.
Both parameterizations roughly maintain full predictive performance until 50\% of the weights have been pruned and decay towards random guessing as more weights are set to zero.
In particular, this pruning strategy does not depend on the variational posterior and is only informed by the compressed weight sample itself, demonstrating that both parameterizations produce compressed samples which are generally capable of maintaining performance to some degree under pruning.

Finally, both parameterizations perform drastically different under KL Divergence pruning.
While Mean-Var test accuracy quickly falls off almost to random guessing after only 50\% of the weights have been set to zero, Mean-KL maintains close to 90\% test accuracy after pruning 90\% of the weights, even outperforming the competitive Absolute Value baseline.
Since this pruning strategy is informed by the variational posterior, the results strongly suggest that, compared to Mean-Var, Mean-KL parameterization leads to a superior variational posterior which produces more robust compressed samples.
Given that this pruning strategy outperforms the competitive baseline, this property is also not a mere peculiarity but could potentially be leveraged to design more robust algorithms.

\section{Conclusion}
\label{sec:conclusion}
We demonstrated that MIRACLE with Mean-KL parameterization bypasses the need for time-consuming KL annealing, leading to training convergence after half the number of optimization steps while maintaining predictive performance.
Furthermore, Mean-KL parameterization produces more meaningful variational posterior distributions with heavy tails, whereas standard Mean-Var parameterization produces distributions which are sharply peaked at particular values.
We illustrated that these qualitative differences result in different properties when exposed to pruning, suggesting that compressed weight samples from Mean-KL are more robust than samples from Mean-Var.
Future work should investigate whether faster convergence properties are scalable to larger models and
pioneer Mean-KL parameterization for Bayesian neural networks independent of compression.
Explicitly utilizing Mean-KL's robustness to design pruning or compression algorithms comprises another possible avenue.

\bibliography{references}

\begin{thebibliography}{13}
\providecommand{\natexlab}[1]{#1}
\providecommand{\url}[1]{\texttt{#1}}
\expandafter\ifx\csname urlstyle\endcsname\relax
  \providecommand{\doi}[1]{doi: #1}\else
  \providecommand{\doi}{doi: \begingroup \urlstyle{rm}\Url}\fi

\bibitem[Blalock et~al.(2020)Blalock, Ortiz, Frankle, and
  Guttag]{blalock2020state}
Blalock, D., Ortiz, J. J.~G., Frankle, J., and Guttag, J.
\newblock What is the {S}tate of {N}eural {N}etwork {P}runing?
\newblock In \emph{Proceedings of Machine Learning and Systems}, 2020.

\bibitem[Chen et~al.(2015)Chen, Wilson, Tyree, Weinberger, and
  Chen]{chen2015compressing}
Chen, W., Wilson, J.~T., Tyree, S., Weinberger, K.~Q., and Chen, Y.
\newblock {C}ompressing {N}eural {N}etworks with the {H}ashing {T}rick.
\newblock In \emph{International Conference on Machine Learning}, 2015.

\bibitem[Corless et~al.(1996)Corless, Gonnet, Hare, Jeffrey, and
  Knuth]{corless1996lambert}
Corless, R.~M., Gonnet, G.~H., Hare, D.~E., Jeffrey, D.~J., and Knuth, D.~E.
\newblock {O}n the {L}ambert {$W$} {F}unction.
\newblock \emph{Advances in Computational Mathematics}, 1996.

\bibitem[Dillon et~al.(2017)Dillon, Langmore, Tran, Brevdo, Vasudevan, Moore,
  Patton, Alemi, Hoffman, and Saurous]{dillon2017tensorflow}
Dillon, J.~V., Langmore, I., Tran, D., Brevdo, E., Vasudevan, S., Moore, D.,
  Patton, B., Alemi, A., Hoffman, M., and Saurous, R.~A.
\newblock {T}ensor{F}low {D}istributions.
\newblock In \emph{arXiv:1711.10604}, 2017.

\bibitem[Flamich(2023)]{flamich2023greedy}
Flamich, G.
\newblock {G}reedy {P}oisson {R}ejection {S}ampling.
\newblock In \emph{arXiv:2305.15313}, 2023.

\bibitem[Flamich et~al.(2020)Flamich, Havasi, and
  Hern{\'a}ndez-Lobato]{flamich2020compressing}
Flamich, G., Havasi, M., and Hern{\'a}ndez-Lobato, J.~M.
\newblock {C}ompressing {I}mages by {E}ncoding their {L}atent {R}epresentations
  with {R}elative {E}ntropy {C}oding.
\newblock In \emph{Advances in Neural Information Processing Systems}, 2020.

\bibitem[Flamich et~al.(2022)Flamich, Markou, and Hernández-Lobato]{Flamich22}
Flamich, G., Markou, S., and Hernández-Lobato, J.~M.
\newblock Fast {R}elative {E}ntropy {C}oding with {A*} {C}oding.
\newblock In \emph{International Conference on Machine Learning}, 2022.

\bibitem[Havasi et~al.(2019)Havasi, Peharz, and Hernández-Lobato]{Havasi19}
Havasi, M., Peharz, R., and Hernández-Lobato, J.~M.
\newblock Minimal {R}andom {C}ode {L}earning: {G}etting {B}its {B}ack from
  {C}ompressed {M}odel {P}arameters.
\newblock In \emph{International Conference on Learning Representations}, 2019.

\bibitem[Higgins et~al.(2017)Higgins, Matthey, Pal, Burgess, Glorot, Botvinick,
  Mohamed, and Lerchner]{Higgins2016betaVAELB}
Higgins, I., Matthey, L., Pal, A., Burgess, C.~P., Glorot, X., Botvinick,
  M.~M., Mohamed, S., and Lerchner, A.
\newblock $\beta$-{VAE}: {L}earning {B}asic {V}isual {C}oncepts with a
  {C}onstrained {V}ariational {F}ramework.
\newblock In \emph{International Conference on Learning Representations}, 2017.

\bibitem[Hinton \& Van~Camp(1993)Hinton and Van~Camp]{hinton1993keeping}
Hinton, G.~E. and Van~Camp, D.
\newblock Keeping {N}eural {N}etworks {S}imple by {M}inimizing the
  {D}escription {L}ength of the {W}eights.
\newblock In \emph{Conference on Computational Learning Theory}, 1993.

\bibitem[Kenton \& Toutanova(2019)Kenton and Toutanova]{kenton2019bert}
Kenton, J. D. M.-W.~C. and Toutanova, L.~K.
\newblock Bert: {P}re-training of {D}eep {B}idirectional {T}ransformers for
  {L}anguage {U}nderstanding.
\newblock In \emph{Conference of the North American Chapter of the Association
  for Computational Linguistics - Human Language Technologies}, 2019.

\bibitem[Paszke et~al.(2019)Paszke, Gross, Massa, Lerer, Bradbury, Chanan,
  Killeen, Lin, Gimelshein, Antiga, Desmaison, Kopf, Yang, DeVito, Raison,
  Tejani, Chilamkurthy, Steiner, Fang, Bai, and Chintala]{PyTorch}
Paszke, A., Gross, S., Massa, F., Lerer, A., Bradbury, J., Chanan, G., Killeen,
  T., Lin, Z., Gimelshein, N., Antiga, L., Desmaison, A., Kopf, A., Yang, E.,
  DeVito, Z., Raison, M., Tejani, A., Chilamkurthy, S., Steiner, B., Fang, L.,
  Bai, J., and Chintala, S.
\newblock {P}y{T}orch: {A}n {I}mperative {S}tyle, {H}igh-{P}erformance {D}eep
  {L}earning {L}ibrary.
\newblock In \emph{Advances in Neural Information Processing Systems}, 2019.

\bibitem[Winitzki(2003)]{Winitzki2003}
Winitzki, S.
\newblock {U}niform {A}pproximations for {T}ranscendental {F}unctions.
\newblock In \emph{Computational Science and Its Applications}, 2003.

\end{thebibliography}
\bibliographystyle{icml2023}

\newpage
\appendix
\onecolumn
\section{KL Divergence Pruning}
\label{app:kl_pruning}
Given a variational posterior $Q_\rvw$ as multivariate Gaussian distribution $\mathcal{N}(\mathbf{w}|\bm{\mu}, \bm{\Sigma})$ with diagonal covariance $\bm{\Sigma} = \mathrm{diag}(\bm{\sigma}^2)$, we want to select the dimension $i$ which minimizes the KL divergence to a Dirac delta centered at zero, that is $\KLD{\delta_\rvw}{Q_\rvw}$.
Because the distribution of $\rvw$ is mean-field factorized, it suffices to consider individual dimensions independ of each other.
To this end, let $Q_{w_i} = \mathcal{N}(w_i|\mu_i, \sigma_i^2)$ and $P_{w_i} = P_w = \mathcal{N}(w|\nu, \rho^2)$, then
\begin{align}
    \KLD{P_w}{Q_{w_i}}
    &= \log \frac{\sigma_i}{\rho} + \frac{\rho^2 + (\nu - \mu_i)^2}{2{\sigma_i}^2} - \frac{1}{2},
\end{align}
which can be simplified if we are only interested in finding the minimizer because $\log \rho$ and $\frac{1}{2}$ are constant with respect to $i$,
\begin{align}
    \underset{i}{\arg \min} \; \KLD{P_w}{Q_{w_i}}
    &= \underset{i}{\arg \min} \; \log \sigma_i + \frac{\rho^2 + (\nu - \mu_i)^2}{2{\sigma_i}^2}.
\end{align}
Now, to let $P_w \to \delta_w$, we first set $\nu = 0$ and let $\rho \to 0$, yielding
\begin{align}
    \underset{i}{\arg \min} \; \KLD{\delta_w}{Q_{w_i}}
    = \underset{i}{\arg \min} \; \log \sigma_i + \frac{{\mu_i}^2}{2{\sigma_i}^2}
    = \underset{i}{\arg \max} \; \log \mathcal{N}(0|\mu_i, \sigma_i),
\end{align}
such that choosing the dimension $i$ by minimizing $\log (\sigma_i) + {\mu_i}^2 / {2\sigma_i}^2$ will prune the weight whose marginal distribution has the lowest KL divergence to a Dirac delta centered at zero or, equivalently, has the highest log density at zero.

\section{Padé Approximation to the Lambert $W$ Function}
\label{app:lambert_w}
Since the Lambert $W$ function, defiend by $W(x) e^{W(x)} = x$, cannot be expressed using elementary functions, it has to be implemented by, for example, numerical or analytical approximations.
We considered three different approximations to the principal branch of the Lambert $W$ function:
Winitzki's approximation for real $x > 0$ (\citealt{Winitzki2003}, (38)), Halley's method for numerical root-finding with cubic rate of convergence, and a Padé approximation of order [3/2].
Winitzki's approximation for real $x > 0$ is used as initialization for Halley's method in the implementation of TensorFlow Probability \citep{dillon2017tensorflow}, however we experienced that the former by itself is not accurate enough and that the latter can be slow and exhibit numerical issues.
Instead, we used a Padé approximation of order [3/2], given by
\begin{align}
    W(x) &\approx
    \frac{
        \frac{13}{720} t(x)^3 + \frac{257}{720} t(x)^2 + \frac{1}{6} t(x) - 1
        }{
        \frac{103}{720} t(x)^2 + \frac{5}{6} t(x) + 1
        }, \\
    \text{where} \qquad
    t(x) &= \sqrt{2 e x + 2},
\end{align}
which was fast and accurate.
We did not consider Winitzki's approximation for $-e^{-1} \leq x \leq 1$ (\citealt{Winitzki2003}, (39)).

\section{Implementation Details}
\label{app:implementation_details}
Our implementation uses PyTorch \citep{PyTorch} and follows \citealt{Havasi19} closely.
The LeNet-5 model consists of two convolutional layers and two linear layers, which are applied sequentially.
The first convolutional layer has 1 input channel, 20 output channels, a kernel size of 5x5, a stride of 1, and no padding.
It is followed by a ReLU activation and a 2D max pooling layer with a kernel size of 2 and a stride of 2.
The second convolutional layer has 20 input channel, 50 output channels, and also a kernel size of 5x5, a stride of 1, and no padding.
It is also followed by a ReLU activation and a 2D max pooling layer with a kernel size of 2 and a stride of 2.
The first linear layer has 800 input features, matching the flattened outputs from the previous layer, 500 output features, and it is followed by a ReLU activation.
The second linear layer has 500 input features and 10 output features, matching the number of classes in the MNIST dataset.
It is followed by a softmax layer to produce class probabilities.
Additionally, weight hashing \citep{chen2015compressing} is used in the second convolutional layer and the first linear layer to reduce the effective number of weights by a factor of 2x and 64x respectively.
The layerwise log standard deviation parameters of the coding distribution were initialized to $-2$.
For Mean-Var parameters, the means were initialized using PyTorch's default initialization and the log standard deviations were initialized to $-10$.
For Mean-KL parameters, $\tau_w$ was initialized by passing PyTorch's default initialization through the analytical inverse of \Cref{eq:mean_kl_mean} and $\gamma_w$ was initialized to $0$.
After initial variational training, we perform 100 fine-tuning steps in-between compressing blocks.
\end{document}